\documentclass[usletter, 10 pt, conference]{URL-ieeeconf}
\IEEEoverridecommandlockouts    % This command is only needed if
\usepackage{graphics}           
\usepackage{times}              
\usepackage{amsmath}            
\usepackage{amssymb}            
\usepackage{graphicx}
\usepackage{algorithm}
\usepackage[noend]{algpseudocode}
\usepackage{booktabs}
\usepackage{color}
\usepackage{multirow}
\usepackage{subcaption}
\usepackage{rotating}
\usepackage{cite}
\usepackage{hyperref}
\hypersetup{
	colorlinks=true,
	linkcolor=black,
	filecolor=black,      
	urlcolor=black,
	pdftitle={Overleaf Example},
	pdfpagemode=FullScreen,
}
\definecolor{instructioncolor}{rgb}{.5,.5,.5}

\def\secref#1{Section~\ref{#1}}
\def\figref#1{Fig.~\ref{#1}}
\def\tabref#1{Table~\ref{#1}}
\def\eqref#1{(\ref{#1})}

\def\vsfig{\vspace{-0.4cm}}

\def\vsequ{\vspace{-0.15cm}}

\newcommand{\rom}[1]{\uppercase\expandafter{\romannumeral #1\relax}}

\makeatletter
\usepackage{xspace}
\DeclareRobustCommand\onedot{\futurelet\@let@token\@onedot}
\def\@onedot{\ifx\@let@token.\else.\null\fi\xspace}

\def\etal{{\textit{et al}}\onedot}
\makeatother

\usepackage{array}
\newcolumntype{L}[1]{>{\raggedright\let\newline\\\arraybackslash\hspace{0pt}}m{#1}}
\newcolumntype{C}[1]{>{\centering\let\newline\\\arraybackslash\hspace{0pt}}m{#1}}
\newcolumntype{R}[1]{>{\raggedleft\let\newline\\\arraybackslash\hspace{0pt}}m{#1}}

\newcommand{\bftab}{\fontseries{b}\selectfont}
\title{\LARGE \bf CHADET: Cross-Hierarchical-Attention for Depth-Completion Using Unsupervised Lightweight Transformer}

\author{Kevin Christiansen Marsim, Jinwoo Jeon, Yeeun Kim, Myeongwoo Jeong, \textit{Member, IEEE}, \\ and Hyun Myung$^{*}$, \textit{Senior Member, IEEE}% <-this % stops a space
	\thanks{$^*$Corresponding author: Hyun Myung}
	\thanks{All the authors are with the School of Electrical Engineering, KAIST (Korea Advanced Institute of Science and Technology), Daejeon, 34141, Republic of Korea. {\tt\scriptsize \{kevinmarsim, zinuok, yeeunk, wjdauddn147, hmyung\}@kaist.ac.kr}} 	 
	\thanks{This work was supported by the Technology Innovation Program~(or Industrial Strategic Technology Development Program-Robot Industry Technology Development)~(00427719, Dexterous and Agile Humanoid Robots for Industrial Applications) and Korea Evaluation Institute of Industrial Technology~(KEIT)~(20018216, Development of mobile intelligence SW for autonomous navigation of legged robots in dynamic and atypical environments for real application) funded by the Ministry of  Trade Industry \& Energy~(MOTIE, Korea). The students are supported by BK21 FOUR.}
}

\begin{document}
\maketitle
 \thispagestyle{empty}
\pagestyle{empty}

%%%%%%%%%%%%%%%%%%%%%%%%%%%%%%%%%%%%%%%%%%%%%%%%%%%%%%%%%%%%%%%%%%%%%%%%%%%%%%%%
\begin{abstract}
  %
  %% WHY  
  Depth information which specifies the distance between objects and current position of the robot is essential for many robot tasks such as navigation. Recently, researchers have proposed depth completion frameworks to provide dense depth maps that offer comprehensive information about the surrounding environment.  
  However, existing methods show significant \mbox{trade-offs} between computational efficiency and accuracy during inference. The substantial memory and computational requirements make them unsuitable for \mbox{real-time} applications, highlighting the need to improve the completeness and accuracy of depth information while improving processing speed to enhance robot performance in various tasks. 
  To address these challenges, in this paper, we propose \textit{CHADET}~({\mbox{cross-hierarchical-attention} \mbox{depth-completion} transformer}), a lightweight \mbox{depth-completion} network that can generate accurate dense depth maps from RGB images and sparse depth points. 
  For each pair, its feature is extracted from the depthwise blocks and passed to the equally lightweight \mbox{transformer-based} decoder. In the decoder, we utilize the novel \mbox{cross-hierarchical-attention} module that refines the image features from the depth information. 
  Our approach improves the quality and reduces memory usage of the depth map prediction, as validated in both KITTI, NYUv2, and VOID datasets.
  %% IMPLEMENTATION, EVALUATION, WHAT FOLLOWS
\end{abstract}
%%%%%%%%%%%%%%%%%%%%%%%%%%%%%%%%%%%%%%%%%%%%%%%%%%%%%%%%%%%%%%%%%%%%%%%%%%%%%%%%
\section{Introduction}
\label{sec:intro}

%%%%%%%%%%%%%%%%%%%
%% WHY: 
% First, answer the WHY question: Why is that relevant? Why should I be
% motivated to read the paper? Why should I care? (1 paragraph, 2-5 sentences)
%Depth completion and estimation techniques have been an active research in robotics community.
Depth maps have been employed in various robotic tasks such as odometry~\cite{karlo2023icra,zuo2021icra}, navigation~\cite{tang2022aaai,biswas2012icra}, and manipulation~\cite{shankar2022ral,schmidt2015icra}. They provide valuable information about the objects surrounding the robot, effectively improving its reaction to environmental changes. The use of sensors to generate depth maps, such as LiDARs, is one of the most widely used methods to provide such information to the robot. 
However, the sparse depth data obtained from a LiDAR sensor often lacks detailed environmental information. Another alternative to the LiDAR sensor is a depth camera which is capable of generating denser depth maps. However, the generated depth maps have a shorter range than LiDAR and are more sensitive to scene illumination\cite{haider2022sensors}.        

%without relying on information from ranging sensors
\begin{figure}[t!]
	\captionsetup{font=footnotesize}
	\centering
	\def\svgwidth{0.4\textwidth}
	\graphicspath{{pics/0826_motivation/}} % important!
	\input{pics/0826_motivation/0826_motivation3.pdf_tex}
	\caption{ Overview of our proposed method. (a) \textit{CHADET} utilizes depthwise blocks and \mbox{cross-hierarchical-attention} to estimate an accurate depth map from an RGB image and sparse depth points. (b) Our proposed \mbox{cross-hierarchical-attention} module combines the hierarchical structure with cross-attention to steadily improve the depth estimation.}
	%USE INKSCAPE!!!! It's a perfect drawing tool for Linux users. Please set \texttt{\textbackslash captionsetup$\{$font$=$footnotesize$\}$} and \texttt{\textbackslash vsfig}. These significantly reduce some spaces and make the paper more luxurious haha~(best viewed in color).}
	\label{fig:motivation2}
	\vspace{-0.7cm}
\end{figure}   
To address these problems,  various depth estimation and completion techniques have been introduced. Depth estimation methods aim to create a complete 3D depth map solely from an RGB image without relying on any existing depth data. 
%Most depth estimation techniques utilize deep learning to generate depth maps from a single~\cite{rudolph2022icra,kusupati2020cvpr} or multiple~\cite{wang2019icra,greene2021icra} RGB images. 
%xu2019iccv
Since sparse depth information is easily accessible through LiDAR or depth sensors, many researchers have attempted to develop depth completion techniques~\cite{lim2020iros} to estimate a complete depth map using sparse depth information alongside additional data from other sensors, such as RGB images.
%from the available incomplete depth information and additional data from other sensors~\cite{lim2020iros,xu2019iccv}, such as RGB images. 
In learning-based depth estimation and completion techniques, convolutional neural networks~(CNN) remain the most popular architecture due to their effective kernel-based feature extraction method~\cite{li2021tnnls}. CNNs show superior performance at capturing local context information but struggle to capture the global context features that model the long-range dependencies between distant features due to their limited kernel size~\cite{vaswani2017attention}.

%due to its ability to capture long-range feature relationships
%, denoted as $\mathrm{Att}$,
%~$\mathbf{Q}$
%~$\mathbf{K}$
%~$\mathbf{V}$
%, computed as follows:
%\begin{equation}
%\mathrm{Att}(\mathbf{Q},\mathbf{K},\mathbf{V}) =\phi\left(\frac{\mathbf{Q}\mathbf{K}^{T}}{\sqrt{d}}\right)\mathbf{V},
%\end{equation}
%where~$\phi$ and~$d$ denote the softmax function and input dimension.
Transformer-based architectures have recently emerged as an alternative to CNNs due to their long-range feature interactions through the attention module~\cite{vaswani2017attention}. Vision transformer, in particular, deals with the global dependency of image features instead of limiting the feature extraction to a local window, improving the model accuracy. 
%Each attention module focuses on different parts of the image and refines those specific parts using values from other patches. 
In the vision transformer architecture, the memory and computation requirement significantly increase due to the construction of query, key, and value for each multi-head attention module~\cite{liu2023cvpr}. This limitation restricts the use of vision transformers in robotic applications which requires~\mbox{real-time} operations.
%However, the transformer architecture is notorious for its highly demanding memory and computation requirements~\cite{liu2023cvpr}.
%due to its complex operations,. 

%The efficient vision transformer is built with channel-wise split attention and reduced channel projection
%, which is an abbreviation of \textit{\mbox{cross-attention} depth-completion efficient transformer}
To address these issues, in this paper, we propose a lightweight depth completion network called ~\mbox{\textit{CHADET}}~({\mbox{cross-hierarchical-attention} depth-completion transformer}) which fuses an RGB image and sparse depth points to predict a dense depth map, as illustrated in~\figref{fig:motivation2}(a). First, {CHADET} extracts features from an RGB image and sparse depth using a depthwise block. Second, a lightweight decoder block is employed to fuse image and depth features through attention mechanism. The layers are designed to compute the features using lightweight components efficiently. Furthermore, unlike the conventional~\mbox{self-attention} approach used in many transformer-based architectures, we introduce a novel \mbox{cross-hierarchical-attention} module to fuse RGB and depth features, as shown in~\figref{fig:motivation2}(b). This module enhances depth estimation by increasing the network depth without increasing the number of network parameters.
%, refining the estimated depth map with the RGB and depth information as shown in~\figref{fig:motivation2}(b). 

%%%%%%%%%%%%%%%%%%%
%% WHICH PROBLEM
% Second, explain WHICH problem you are solving/address to solve.

%%%%
%%%%%%%%%%%%%%%%%%%
%% HOW & WHAT
% Third, explain briefly how one can address the problem in general and mention 
% briefly what others/we before have done. Prepare the reader for your contribution 
% that comes in the next section (and not here!).

%%%%%%%%%%%%%%%%%%%
%% MAIN CONTRIBUTION & WHAT FOLLOWS FROM THAT
% Explain your contribution in one paragraph. This is a very important paragraph. 
% Always start that paragraph with: ``The main contribution of this paper is''

%The main contribution of this paper is a \dots  
The main contributions of this paper are threefold:
\begin{enumerate}
	%\item We propose a real-time cooperative localization framework utilizing low-cost sensors to estimate the state of each robot simultaneously.
	\item We propose a lightweight depth completion network that predicts dense depth maps with reduced computational load without compromising depth estimation performance.
	\item We propose a \mbox{cross-hierarchical-attention} transformer decoder block that allows RGB feature refinement from the depth feature, resulting in accurate depth maps with clear distinctions between objects and their background.
	\item We validate our network on the KITTI, NYUv2, and VOID datasets to demonstrate its performance across various indoor and outdoor environments.
\end{enumerate}

%%%%%%%%%%%%%%%%%%%
%% OUR KEY CLAIMS (can be merged with the main contribution above if desired)
% Explicitly(!) state your claims in one (short) paragraph and make
% sure you pick them up again in the experiments and support every claim.

%In sum, we make three key claims:
%Our approach is able to
%
%(i) \dots;
%
%(ii) \dots;
%
%(iii) \dots.
%
%These claims are backed up by the paper and our experimental evaluation.

%%%%%%%%%%%%%%%%%%%%%%%%%%%%%%%%%%%%%%%%%%%%%%%%%%%%%%%%%%%%%%%%%%%%%%%%%%%%%%%%
\section{Related Works}
\label{sec:related}

% Discuss the main related work and cite around 15-25 papers in sum. 
% The related work section should be approx. 1 column long, assuming 
% a 6-page paper.  Structure the section in paragraphs, grouping the 
% papers, and describing the key approaches with 1-2 sentences. If 
% applicable, describe the key difference to your approach at the end 
% of each paragraph briefly. Avoid adding subsections, al least for a 
% conference paper.

%Lee~\textit{et al.}~\cite{lee2019icra} designed a depth-completion network consisting of geometry and context networks. The geometry network learns the initial depth map and its estimated surface normal, while the context network tries to refine the initial depth map from the sparse depth.
\subsection{Depth Completion}
The depth completion field combines sparse depth measurements with other input modalities such as RGB or semantic images~\cite{jaritz2018threedv}. In learning-based depth completion, the supervised depth completion networks learn to estimate the dense depth map by directly minimizing the error between the ground truth and its output. Hu~\textit{et al.}~\cite{hu2021icra} proposed PENet, a dual-branch network combining an RGB image and LiDAR point cloud through a color-dominant and depth-dominant branch. 
%Lee~\textit{et al.}~\cite{lee2019icra} designed a depth-completion network consisting of geometry and context networks. 
Lim~\textit{et al.}~\cite{lim2020iros} proposed a multi-stage depth completion network that stacks multiple encoder-decoder networks to complete a depth map from an RGB image and a 2D LiDAR measurement. 
%Choi~\textit{et al.}~\cite{choi2021icra} estimated the depth map using three-stage networks that extrapolate the RGB image into a stereo pair for more accurate refinement in later stages. 
These methods are capable of producing an accurate depth map in most datasets. However, the reliance on ground truth data for training limits their applicability in many robotic tasks, as acquiring such data in real-world scenarios can be challenging.  

Unsupervised learning methods have been developed to bypass the ground truth requirement for depth completion. 
%These methods try to minimize other loss functions, such as photometric error, to generate guidance cues that can be derived from collected sensor data. 
%Ma~\textit{et al.}~\cite{ma2019icra} developed a self-supervised training framework that optimizes sparse depth, photometric, and smoothness errors to produce a smooth depth map from RGB images and a sparse LiDAR point cloud. Fan~\textit{et al.}~\cite{fan2022icra_depth} proposed two unsupervised depth completion networks, which are completion and denoising networks, for an RGB-D camera.
%The completion network takes RGB and depth images to predict the affinity matrix and propagates it through the spatial propagation method to reduce blurred areas in the depth map. 
Wong~\textit{et al.}~\cite{wong2020ral} proposed unsupervised depth completion networks that utilize visual feature points from a visual odometry system. The subsequent work includes learning topological features from synthetic data~\cite{wong2021ral} and incorporating camera's intrinsic parameters into the network~\cite{wong2021iccv}. These methods can effectively produce dense depth maps without requiring any ground truth information. Jeon~\textit{et al.}~\cite{jeon2022ral} developed a novel unsupervised depth completion network that utilizes line features from visual odometry. However, their outputs lack the distinction between background and objects which improves the completion accuracy. 

%Several researchers have utilized vision transformer architecture for depth completion tasks due to its capability of capturing the relationship between feature patches.

\subsection{Transformer-based Depth Completion}
Vision transformer architectures have been extensively utilized for depth completion tasks due to their ability to capture relationships between feature patches. Qian~\mbox{\textit{et al.}~\cite{qian2022iarce}} proposed SDFormer with a U-shaped \mbox{encoder-decoder} \mbox{self-attention} transformer architecture to extract features at different levels. 
%SDFormer employs a self-attention mechanism by concatenating the RGB image features with the depth features. 
Zhang~\textit{et al.}~\cite{zhang2023cvpr} proposed a joint convolution and transformer module to extract local and global context information in a single layer for depth completion. 
%This hybrid module aims to extract local and global context information of the features coming from the projection of RGB and depth features. 
Rho~\textit{et al.}~\cite{rho2022cvpr} extended PENet~\cite{hu2021icra} by introducing \mbox{self-attention} and \mbox{guided-attention} modules into the network. The \mbox{guided-attention} module utilizes \mbox{multi-modal} information fusion but it only serves as a shortcut connection between modules. Jia~\textit{et al.}~\cite{jia2023arxiv} proposed a gated cross-attention network for depth completion via a gating mechanism that produces confidence features. A ray tune mechanism is also applied to determine the optimal number of training iterations.

Transformer-based depth completion networks are capable of capturing long-range dependencies between features. However, their large memory requirement and long processing time limit their usage in many robotic applications. Therefore, research on lightweight \mbox{transformer-based} depth completion networks is essential to ensure accuracy and~\mbox{real-time} performance.

%% BRIEFLY SUMMARIZE OWN CONTRIBUTION 

%%%%%%%%%%%%%%%%%%%%%%%%%%%%%%%%%%%%%%%%%%%%%%%%%%%%%%%%%%%%%%%%%%%%%%%%%%%%%%%%
%\begin{figure*}[t!]
%	\captionsetup{font=footnotesize}
%	\centering
%	\def\svgwidth{0.95\textwidth}
%	\graphicspath{{pics/1218_framework/}} % important!
%	\input{pics/1218_framework/1218_framework.pdf_tex}
%	\caption{Overall framework of our algorithm. We utilize the VIO algorithm to obtain the ego motion of the robot and couple it with UWB measurement to correct the trajectory. First, UWB-based triangulation is performed to obtain a triangulated pose based on other robots' VIO poses and UWB data $d^{i,k}_t$. The resulting triangulation result $\mathrm{\hat{\bf{T}}}^{w,k}_t$ is stored in the sliding window. Second, error-based optimization uses the triangulated pose within the sliding window to estimate the VIO drift $\Delta\mathrm{\hat{\bf{T}}}^{w,k}_t$.}
%	%USE INKSCAPE!!!! It's a perfect drawing tool for Linux users. Please set \texttt{\textbackslash captionsetup$\{$font$=$footnotesize$\}$} and \texttt{\textbackslash vsfig}. These significantly reduce some spaces and make the paper more luxurious haha~(best viewed in color).}
%	\label{fig:framework}
%	\vsfig
%\end{figure*}

\section{CHADET}
\label{sec:main}
%% Describe your approach. It is okay to divide the main section
%%  into a few subsections (e.g., 2-4 subsections).
\subsection{Overall Framework}
\begin{figure*}[t!]
	\captionsetup{font=footnotesize}
	\centering
	\def\svgwidth{0.98\textwidth}
	\graphicspath{{pics/0723_network/}} % important!
	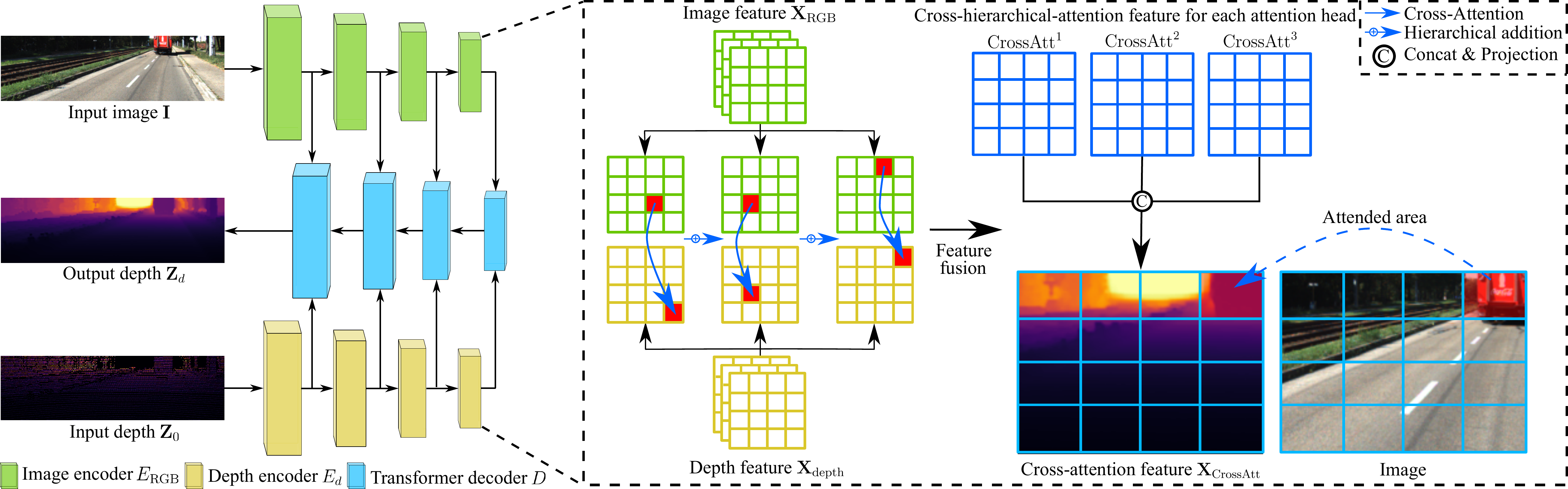
	\caption{{CHADET} takes an RGB image $\mathbf{I}$ and sparse depth points $\mathbf{Z}_{0}$ as inputs to produce output depth $\mathbf{Z}_{\textrm{d}}$. Each encoder ($E_{\textrm{{RGB}}}$ and $E_{\textrm{{depth}}}$) layer encodes~$\mathbf{I}$ and~$\mathbf{Z}_{0}$ into features utilized in the next stage and decoder branch. The dashed line illustrates detailed operation at each transformer decoder layer~$D$. The transformer decoder attempts to selectively refine the current features utilizing the~$\mathbf{X}_{\textrm{RGB}}$ and~$\mathbf{X}_{\textrm{depth}}$, which are image and depth features, respectively, via \mbox{cross-hierarchical-attention}. The~$\mathbf{X}_{\textrm{RGB}}$ and~$\mathbf{X}_{\textrm{depth}}$ are split into multiple channels in the \mbox{cross-hierarchical-attention} module. Each channel focuses on different windows corresponding to similar scenes between the image and depth features, resulting in~$\mathrm{CrossAtt}^i$ at the~$i$-th head. Then, the attended feature from the previous head $\mathrm{CrossAtt}^{i-1}$ is added to the current head. Finally, all~$\mathrm{CrossAtt}^i$ are combined via concatenation and projection to produce the cross-attention feature~$\mathbf{X}_{\textrm{CrossAtt}}$.  %Finally, the results of each \mbox{cross-attention} module are fused through the concatenation and projection to produce the estimated dense depth at each scale.
	}
	%USE INKSCAPE!!!! It's a perfect drawing tool for Linux users. Please set \texttt{\textbackslash captionsetup$\{$font$=$footnotesize$\}$} and \texttt{\textbackslash vsfig}. These significantly reduce some spaces and make the paper more luxurious haha~(best viewed in color).}
	\label{fig:network}
	\vsfig
\end{figure*}
%The overall objective of our method is to estimate a full depth map given RGB image and sparse depth inputs. 
This section explains the proposed network, {CHADET}, which estimates a complete depth map from an RGB image and sparse depth points. The overall framework of {CHADET} is shown in~\figref{fig:network}. The proposed network consists of an image encoder, a depth encoder, and a \mbox{transformer-based} decoder block. %The U-shaped network independently extracts features from RGB images and depth features. 
A depthwise block is proposed to replace the original encoder block for smaller memory consumption. Both image and depth encoders adopt the same depthwise block architecture. After obtaining the embedding features, they are passed onto the proposed \mbox{transformer-based} decoder block. The proposed transformer-decoder block is utilized to fuse multimodal information with low memory usage while producing accurate depth prediction. At every stage, the output of the previous decoder is refined by the proposed \mbox{cross-hierarchical-attention} module to obtain the depth map details. The algorithm is formulated as follows:
\vsequ
\begin{equation}
\begin{aligned}
	\mathbf{Z}_d=D\left(E_{\textrm{{RGB}}}\left(\mathbf{I}\right),E_{\textrm{\em{d}}}\left(f_\textrm{sd}\left(\mathbf{Z}_{0}\right)\right)\right), 
\end{aligned}
\vsequ
\end{equation}
where~$\mathbf{I}$,~$\mathbf{Z}_0$, and~$\mathbf{Z}_d$ denote the input RGB image, sparse depth, and output dense depth map;~$E_\textrm{RGB}\left(\cdot\right)$,~$E_d\left(\cdot\right)$,~$f_\textrm{sd}\left(\cdot\right)$, and~$D\left(\cdot\right)$ represent the image encoder, depth encoder, the sparse-to-dense module, and \mbox{transformer-based} decoder, respectively.

\subsection{RGB and Sparse Depth Encoder}
\label{subsec:network_arch}
\begin{figure}[t!]
	\captionsetup{font=footnotesize}
	\centering
	\def\svgwidth{0.45\textwidth}
	\graphicspath{{pics/0724_Encoder/}} % important!
	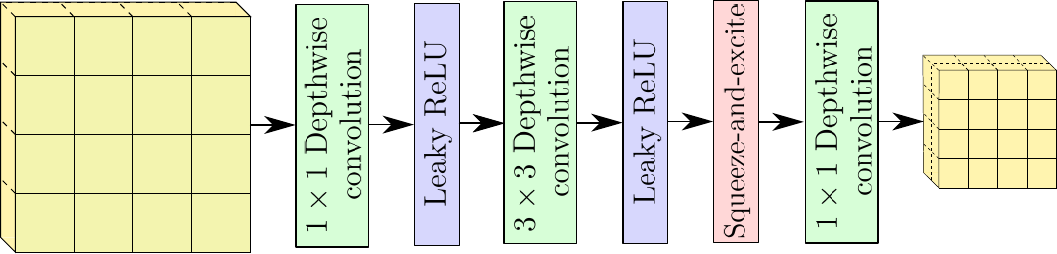
	\caption{The encoder layer consists of depthwise convolution and \mbox{squeeze-and-excite}~\cite{hu2018cvpr} modules. Depthwise convolution efficiently extracts the feature from the input with one filter per channel. Squeeze-and-excite module improves the channel interaction to compensate for the previous depthwise feature extraction.}
	%USE INKSCAPE!!!! It's a perfect drawing tool for Linux users. Please set \texttt{\textbackslash captionsetup$\{$font$=$footnotesize$\}$} and \texttt{\textbackslash vsfig}. These significantly reduce some spaces and make the paper more luxurious haha~(best viewed in color).}
	\label{fig:encoder}
	\vspace{-0.6cm}
\end{figure}
In the early stages of the network, RGB and depth features are extracted by utilizing~$E_{\textrm{{RGB}}}\left(\cdot\right)$ and~$E_\textrm{\em{d}}\left(\cdot\right)$, respectively, sharing the same structure as shown in~\figref{fig:encoder}. The encoder blocks are expressed as follows:
\vsequ
\begin{equation}
\begin{aligned}
\mathbf{X}_{\textrm{depth}}&=E_{\textrm{\em{d}}}\left(f_\textrm{sd}\left(\mathbf{Z}_0\right)\right), \\ \mathbf{X}_{\textrm{RGB}}&=E_{\textrm{{RGB}}}\left(\mathbf{I}\right),
\end{aligned}
\vsequ
\end{equation}
where~$\mathbf{X}_{\textrm{{depth}}}$ and~$\mathbf{X}_{\textrm{{RGB}}}$ denote the depth and RGB features, respectively. The RGB image is directly fed into the depthwise block, consisting of depthwise convolutional layers~\cite{sandler2018cvpr} and a squeeze-and-excite~\cite{hu2018cvpr} module. The depthwise convolutional layer extracts local image features while reducing memory usage. To enhance the representational quality of channel embeddings, a squeeze-and-excite module is appended to the encoder block, leveraging global pooling and a fully connected layer, crucial for the lightweight transformer decoder discussed in~\secref{subsec:efficient_trans}.

For the sparse depth input, we first utilize the \mbox{sparse-to-dense} module introduced by~\cite{wong2021iccv} to obtain an initial \mbox{quasi-dense} depth map. 
%The sparse-to-dense module consists of several multi-scale min and max-pooling operations followed by convolutional layers to combine the multi-scale depth features into the initial depth map.
The initial depth map is utilized as input to the depthwise block. 
%The depth encoder block consists of the same components to the image encoder block. We additionally extract the estimated dense depth for every scale that will be used further in the decoder block,

\subsection{Lightweight Transformer-based Decoder}
\label{subsec:efficient_trans}
\begin{figure}[t!]
	\captionsetup{font=footnotesize}
	\centering
	\def\svgwidth{0.45\textwidth}
	\graphicspath{{pics/0724_Decoder/}} % important!
	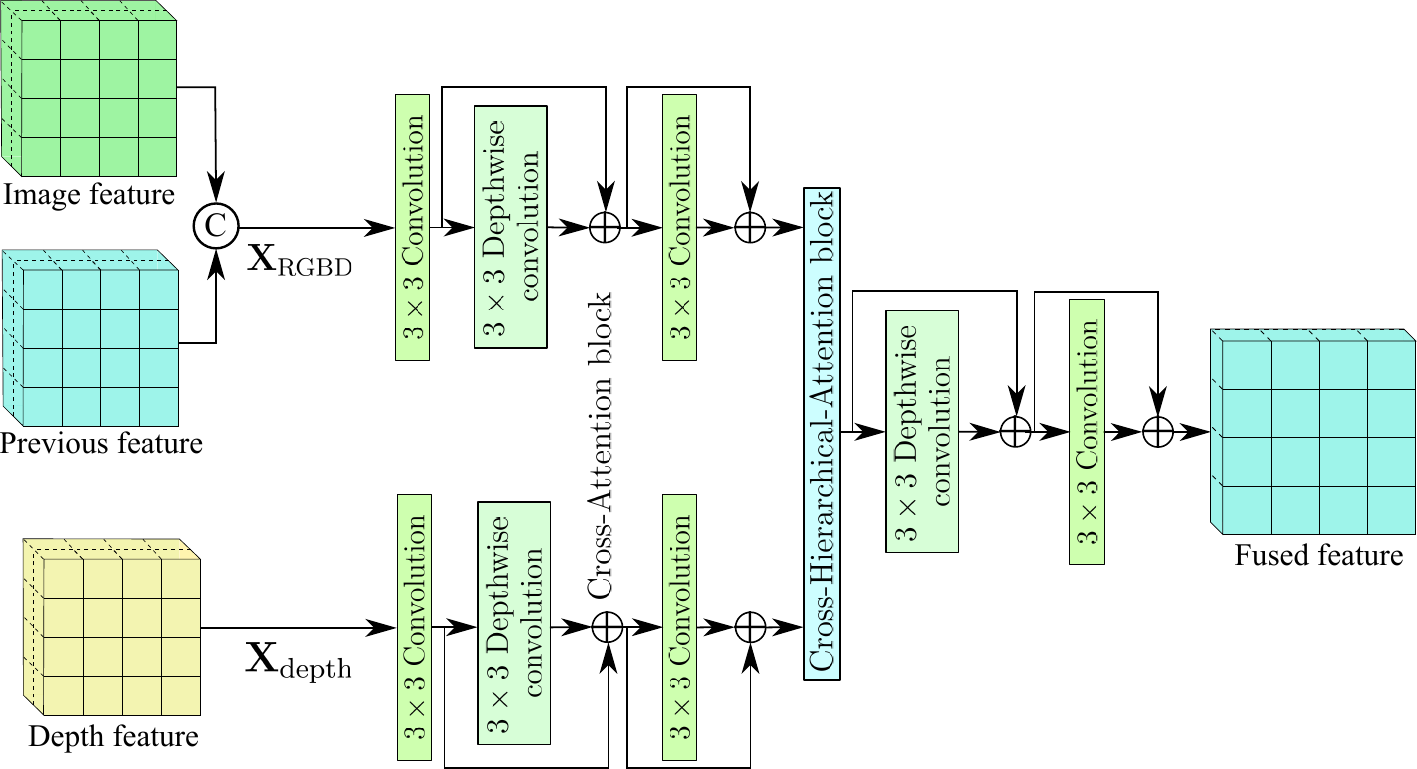
	\caption{The decoder layer with a lightweight \mbox{cross-hierarchical-attention} module to combine both image and depth features. First, the image feature is combined with the feature from a previous decoder layer through concatenation and 3$\times$3 convolution to form the feature $\mathbf{X}_{\textrm{RGBD}}$. Next, $\mathbf{X}_{\textrm{RGBD}}$ and $\mathbf{X}_{\textrm{depth}}$ are further processed through residual correction of 3$\times$3 depthwise convolution and 3$\times$3 convolution. Finally, the \mbox{cross-hierarchical-attention} module is utilized to combine both features in order to obtain the final decoder feature.}
	%USE INKSCAPE!!!! It's a perfect drawing tool for Linux users. Please set \texttt{\textbackslash captionsetup$\{$font$=$footnotesize$\}$} and \texttt{\textbackslash vsfig}. These significantly reduce some spaces and make the paper more luxurious haha~(best viewed in color).}
	\label{fig:decoder}
	\vspace{-0.2cm}
\end{figure}

We propose a novel \mbox{transformer-based} decoder block that efficiently and effectively fuses information from the RGB features, estimated dense depth, and depth features, as shown in~\figref{fig:decoder}. 
%Our decoder block can be formulated as follows:
%\vsequ
%\begin{equation}
%\mathbf{Z}_{d}=D\left(\mathbf{X}_\textrm{RGB},\mathbf{X}_\textrm{depth}\right).
%\vsequ
%\end{equation} 
Let the feature generated from the previous decoder block be denoted as~$\mathbf{Z}_{k-1}$. If it is the first decoder layer,~$\mathbf{Z}_{k-1}$ is a null matrix with the same dimension as~$\mathbf{X}_{\textrm{{RGB}}}$. First, the feature~$\mathbf{X}_{\textrm{RGBD}}$ is constructed by concatenating $\mathbf{X}_{\textrm{RGB}}$ and $\mathbf{Z}_{k-1}$ followed by a 3$\times$3 convolution. It effectively fuses the RGB feature with its estimated depth projection based on the learned features. Then,~$\mathbf{X}_\textrm{RGBD}$ and~$\mathbf{X}_\textrm{depth}$ are refined via the concatenation of depthwise convolutional and regular convolutional layers with residual connection.
%defined as follows:
%\begin{equation}
%\begin{aligned}
%&\mathbf{X}_{RGB3D}=\omega\left(\mathbf{X}_{RGB} \ %\textcircled{c} \ \mathbf{Z}_{k-1}\right),\\
%&\mathbf{X}_{RGB3D}=\varpi\left(\mathbf{X}_{RGB3D}\right)+\mathbf{X}_{RGB3D},\\
%&\mathbf{X}_{RGB3D}=\omega\left(\mathbf{X}_{RGB3D}\right)+\mathbf{X}_{RGB3D},
%\end{aligned}
%\end{equation}
%\begin{equation}
%\begin{aligned}
%&\mathbf{X}_{depth}=\omega\left(\mathbf{X}_{depth}\right),\\
%&\mathbf{X}_{depth}=\varpi\left(\mathbf{X}_{depth}\right)+\mathbf{X}_{depth},\\
%&\mathbf{X}_{depth}=\omega\left(\mathbf{X}_{depth}\right)+\mathbf{X}_{depth},
%\end{aligned}
%\end{equation}
%where~$\omega$,~$\varpi$, and~$\textcircled{c}$ denote the convolution, depthwise convolution, and concatenation operations, respectively. 
%Next, 
The~$\mathbf{X}_{\textrm{RGBD}}$ and~$\mathbf{X}_\textrm{depth}$ are combined by the transformer block, as shown in~\figref{fig:cross_att}. However, it has been shown that the attention module inefficiently exploits high memory usage due to multiple overlapping information between channels~\cite{liu2023cvpr}. To tackle this problem, we propose the cross-hierarchical-attention architecture,  inspired by~\cite{zhang2022cvpr}, to reduce the computation overhead of the multi-head attention. To achieve this, the features are divided channel-wise into~\mbox{$n$-heads} as follows:
\begin{equation}
\begin{aligned}
\mathrm{split}(\mathbf{X}_{\textrm{RGBD}}) &= \left\{\mathbf{X}^1_{\textrm{RGBD}},\mathbf{X}^2_{\textrm{RGBD}},\dots,\mathbf{X}^n_{\textrm{RGBD}}\right\},\\
\mathrm{split}(\mathbf{X}_{\textrm{depth}})&= \left\{\mathbf{X}^1_{\textrm{depth}},\mathbf{X}^2_{\textrm{depth}},\dots,\mathbf{X}^n_{\textrm{depth}}\right\},
\end{aligned}
\end{equation}
where~$\mathrm{split}$ denotes the channel-wise split function;~$\mathbf{X}^i_{\textrm{\textrm{RGBD}}}$ and~$\mathbf{X}^i_{\textrm{{depth}}}$ denote the RGBD and depth features at the $i$-th head~($i$~=~1,~\dots,~$n$), respectively. In order to further reduce the computation requirements, the channel information is compressed into a smaller size via the query~\mbox{$\mathbf{W}_{Q} \in \mathbb{R}^{d_{c}\times d_{q}}$}, key~\mbox{$\mathbf{W}_{K} \in \mathbb{R}^{d_{c}\times d_{k}}$}, and value~\mbox{$\mathbf{W}_{V} \in \mathbb{R}^{d_{c}\times d_{v}}$} projection weights, where~$d_c$,~$d_q$,~$d_k$, and~$d_v$ denote the number of the original feature, query projection, key projection, and value projection channels, respectively.~$d_q$,~$d_k$, and~$d_v$ are kept to be significantly smaller than~$d_c$. Therefore, the query, key, and value with reduced channel sizes are obtained while retaining the required information for attention computation. 
Subsequently, the attention is computed for each head using their respective query~$\mathbf{Q}$, key~$\mathbf{K}$, and value~$\mathbf{V}$.
 %The result of each head will be combined via concatenation and projection to produce the output feature.   

\subsection{Cross-hierarchical-attention Module}
\label{subsec:cross_att}
\begin{figure}[t!]
	\captionsetup{font=footnotesize}
	\centering
	\def\svgwidth{0.45\textwidth}
	\graphicspath{{pics/0723_cross_attention/}} % important!
	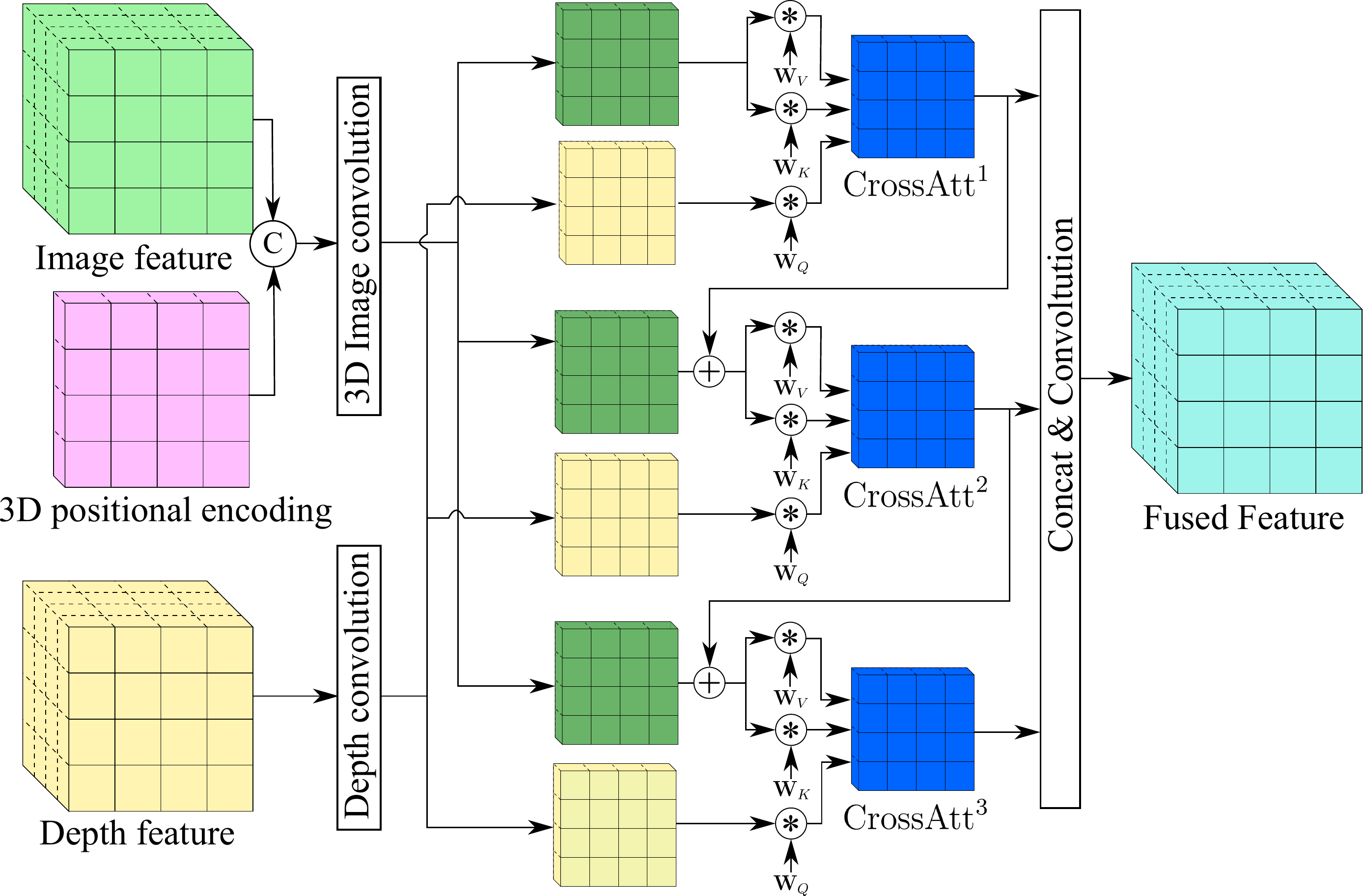
	\caption{Overview of our \mbox{cross-hierarchical attention} module design. The \mbox{cross-attention} $\mathrm{CrossAtt}^i$ calculation utilizes different sources of query, key, and value at the $i$-th head by mutiplying it with projection weights $\mathbf{W}_Q$, $\mathbf{W}_K$, and $\mathbf{W}_V$. First, channel-wise split is performed to divide the features into several heads. Next, the depth feature is utilized as query and the image feature as key-value. For each attention head, the previous attention result is utilized in the current attention head by directly adding it with the image feature. This assignment allows detailed correction of the estimated features with \mbox{inter-sensor} data through attention mechanism.}
	%First, we do channel-wise split of the features into several heads.
	%USE INKSCAPE!!!! It's a perfect drawing tool for Linux users. Please set \texttt{\textbackslash captionsetup$\{$font$=$footnotesize$\}$} and \texttt{\textbackslash vsfig}. These significantly reduce some spaces and make the paper more luxurious haha~(best viewed in color).}
	\label{fig:cross_att}
	\vspace{-0.6cm}
\end{figure}
In the \mbox{self-attention} mechanism, the features are refined by other patches in the same window according to the attention equation, denoted as $\mathrm{Att}$, as follows:
\begin{equation}
\mathrm{Att}(\mathbf{Q},\mathbf{K},\mathbf{V}) =\phi\left(\frac{\mathbf{Q}\mathbf{K}^{T}}{\sqrt{d}}\right)\mathbf{V},
\end{equation}
where~$\phi$ and~$d$ denote the softmax function and input dimension. However, the resulting features are limited only to their local window without considering the other sensor information. In contrast, the \mbox{cross-attention} mechanism utilizes the other sensor information to compute the attention scores and refine the features. This method effectively fuses the long-range dependencies between the RGB and depth features, even when they are separated by objects such as occlusions. Therefore, our module adopts the \mbox{cross-attention} mechanism to enhance feature fusion between RGB and depth modalities.
%that are separated by other objects, such as occlusion.

\figref{fig:cross_att} illustrates the \mbox{cross-hierarchical-attention} module utilized in our network. The depth feature is utilized as query~$\mathbf{Q}^i_{\textrm{{depth}}}$ while the RGBD feature is used as the key~$\mathbf{K}^i_{\textrm{{RGBD}}}$ and value~$\mathbf{V}^i_{\textrm{{RGBD}}}$ in the~$i$-th head of the transformer. We utilize the depth as query as it contains the depth cues describing the actual desired depth map. The~$\mathbf{X}_{\textrm{{RGBD}}}$ is combined with the 3D positional encoding, as mentioned in~\cite{wong2021iccv}, to encode the spatial relationship of the features. The attention score is calculated based on the similarity between~$\mathbf{Q}^i_{\textrm{{depth}}}$ and~$\mathbf{K}^i_\textrm{RGBD}$. Then, the~$\mathbf{V}^i_\textrm{RGBD}$ is multiplied by the attention scores to identify the most prominent depth feature and refine the RGBD features. The resulting features are used to estimate the dense depth output of the current head. 

In another work~\cite{zhang2022cvpr}, each attention head results are combined directly via concatenation and projection. However, this late fusion strategy may lead to the loss of \mbox{fine-grained} details produced by individual attention heads. It treats each head's output independently without considering their sequential dependencies. To address this limitation, we propose a simple yet effective hierarchical attention strategy, which leverages the output of the previous attention head $\mathrm{CrossAtt}^{i-1}$ as input for the current attention head. Specifically, rather than independently processing each attention head, our approach directly integrates the accumulated information by adding the current image feature $\mathbf{X}_{\textrm{{RGBD}}}^i$ with $\mathrm{CrossAtt}^{i-1}$. This ensures a progressive refinement of features across multiple attention heads, as illustrated in~\figref{fig:cross_att}. By structuring our attention mechanism hierarchically, information flows smoothly from the first attended feature to the latest one, enabling a more comprehensive and context-aware fusion of RGB and depth cues. The \mbox{cross-hierarchical-attention} process at the $i$-th head, denoted as $\mathrm{CrossAtt}^i\left(\cdot\right)$, is expressed as follows:
%\begin{equation}
%\begin{aligned}
%& \ \ \  \ \ \ \ \ \ \ \ \ \ \mathbf{Q}^i_{\textrm{RGB3D}}=  \mathbf{W}_Q * \mathbf{X}^i_{\textrm{RGB3D}},\\
%&\ \ \ \ \  \ \ \ \ \ \ \ \ \ \ \mathbf{K}^i_{\textrm{depth}}= \mathbf{W}_K * \mathbf{X}^i_{\textrm{depth}}, \\
%& \ \ \ \ \  \ \ \ \ \ \ \ \ \ \ \mathbf{V}^i_{\textrm{depth}}= \mathbf{W}_V * \mathbf{X}^i_{\textrm{depth}}, \\
%&\mathrm{CrossAtt}^i(\mathbf{Q}^i_{\textrm{RGB3D}},\mathbf{K}^i_{\textrm{depth}},\mathbf{V}^i_{\textrm{depth}})= \\ & \ \  \ \ \  \ \ \ \ \  \ \  \ \ \  \ \ \  \ \ \ \ \mathrm{Att}(\mathbf{Q}^i_{\textrm{RGB3D}},\mathbf{K}^i_{\textrm{depth}},\mathbf{V}^i_{\textrm{depth}}),
%\end{aligned}
%\end{equation}
\begin{equation}
\begin{aligned}
\mathbf{Q}^i_{\textrm{depth}}&=  \mathbf{W}_Q * \mathbf{X}^i_{\textrm{depth}},\\
\mathbf{K}^i_{\textrm{RGBD}}&= \mathbf{W}_K * \left(\mathbf{X}^i_{\textrm{RGBD}}+\mathrm{CrossAtt}^{i-1}\right), \\ \mathbf{V}^i_{\textrm{RGBD}}&= \mathbf{W}_V * \left(\mathbf{X}^i_{\textrm{RGBD}}+\mathrm{CrossAtt}^{i-1}\right), \\
\mathrm{CrossAtt}^i&=\mathrm{Att}(\mathbf{Q}^i_{\textrm{depth}},\mathbf{K}^i_{\textrm{RGB3D}},\mathbf{V}^i_{\textrm{RGB3D}}),
\end{aligned}
\vsequ
\end{equation}
%\begin{equation}
%\begin{aligned}
%
%\end{aligned}
%\end{equation}
%\mathrm{CrossAtt}^i(\mathbf{Q}^i_{\textrm{depth}},\mathbf{K}^i_{\textrm{RGB3D}},\mathbf{V}^i_{\textrm{RGB3D}})= \\ & \ \  \ \ \  \ \ \ \ \  \ \  \ \ \  \ \ \  \ \ \ \ \mathrm{Att}(\mathbf{Q}^i_{\textrm{depth}},\mathbf{K}^i_{\textrm{RGB3D}},\mathbf{V}^i_{\textrm{RGB3D}}),
where $*$ denotes inner product operation. In the next layer, the results of each head are combined by concatenating and projecting them to produce the final estimated inverse dense depth~$\mathbf{Z}_{k}$, which is defined as follows:
\begin{equation}
\begin{aligned}
\mathbf{X}_{\textrm{CrossAtt}}&=\mathrm{CrossAtt}^1 \ \textcircled{c} \ \mathrm{CrossAtt}^2 \ \textcircled{c} \ \dots \ \textcircled{c} \ \mathrm{CrossAtt}^n, \\
\mathbf{Z}_{k}&=\omega\left(\mathbf{X}_\textrm{CrossAtt}\right),
\end{aligned}
\end{equation}
where $\mathbf{X}_{\textrm{CrossAtt}}$ denotes the concatenated feature from all attention heads; $\omega$ and~$\textcircled{c}$ denote convolution and concatenation operations, respectively.

%\begin{figure}[t!]
%	\captionsetup{font=footnotesize}
%	\centering
%	\def\svgwidth{0.48\textwidth}
%	\graphicspath{{pics/0725_upprojection/}} % important!
%	\input{pics/0725_upprojection/0725_upprojection2.pdf_tex}
%	\caption{The upsample layer with residual and transposed convolutions to restore the decoder feature size to the original image size. The squeeze-and-excite module is utilized again to improve the decoder feature channel interaction.}
%	%USE INKSCAPE!!!! It's a perfect drawing tool for Linux users. Please set \texttt{\textbackslash captionsetup$\{$font$=$footnotesize$\}$} and \texttt{\textbackslash vsfig}. These significantly reduce some spaces and make the paper more luxurious haha~(best viewed in color).}
%	\label{fig:upproject}
%	\vspace{-0.6cm}
%\end{figure}
%, as shown in~\figref{fig:upproject}, 
\begin{figure}[t!]
	\captionsetup{font=footnotesize}
	\centering
	\def\svgwidth{0.48\textwidth}
	\graphicspath{{pics/0725_upprojection/}} % important!
	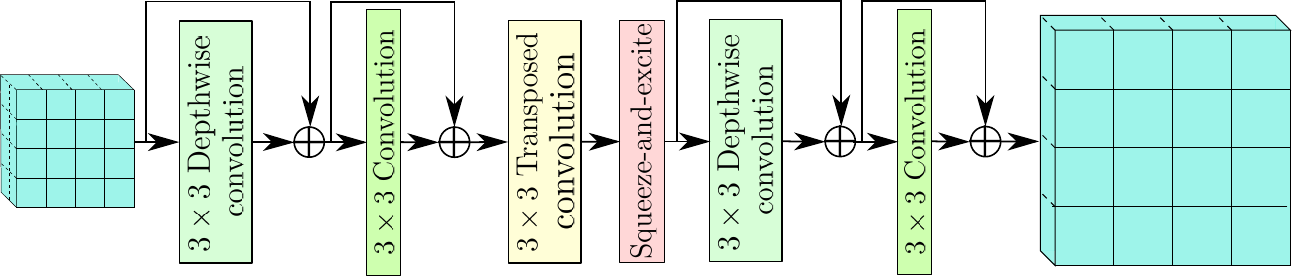
	\caption{The upsample layer with residual and transposed convolutions to restore the decoder feature size to the original image size. The squeeze excite module is utilized again to improve the decoder feature channel interaction.}
	%USE INKSCAPE!!!! It's a perfect drawing tool for Linux users. Please set \texttt{\textbackslash captionsetup$\{$font$=$footnotesize$\}$} and \texttt{\textbackslash vsfig}. These significantly reduce some spaces and make the paper more luxurious haha~(best viewed in color).}
	\label{fig:upproject}
	\vspace{-0.6cm}
\end{figure}

Between each decoder block,~$\mathbf{Z}_k$ is upsampled by the upsample layer to match the size of the next stage features while still keeping information from the smaller window as shown in~\figref{fig:upproject}. The first two operations of the upsample layer are the depthwise convolutional and regular convolutional layers with a residual connection from the encoder. They are followed by 3$\times$3 transposed convolutional layer and squeeze-and-excite module to increase the feature size while maintaining the quality of channel embeddings. Finally, the first two operations are repeated to fuse information between channels.

\subsection{Depth Estimation}
The final decoder layer outputs a normalized inverse depth map from the sigmoid activation function. The estimated dense depth map is computed as follows:
\begin{equation}
\mathbf{Z}_d= \frac{\textrm{min}_d}{\sigma(\mathbf{Z}_k)+\frac{\textrm{min}_d}{\textrm{max}_d}},
\end{equation}
where~$\sigma$ denotes the sigmoid function;~$\textrm{min}_d$ and~$\textrm{max}_d$ denote the estimated minimum and maximum depth values, respectively.

\subsection{Loss Function}
\label{subsec:loss}

We employ the unsupervised learning framework for its superior adaptability to various environments where obtaining ground truth is not feasible. The objective of the network is to minimize three loss functions, which are defined as follows:
\begin{equation}
L=w_p l_p+ w_d l_d + w_s l_s,
\vsequ
\end{equation}
where~$l_p$,~$l_d$, and~$l_s$ denote the photometric loss, sparse depth loss, and local smoothness loss;~$w_p$,~$w_d$, and~$w_s$ denote the weight of each loss, respectively.

%In the training stage, we collect consecutive RGB images~\mbox{$\mathbf{I}_\tau$}, where $\tau\in{\{t-1, \ t,\ t+1\}}$; $t$ denotes the point cloud acquisition time. 
%and $\mathbf{Z}_0$ and its sparse depth measurement 

The photometric loss measures the error between pixel intensities or color values between a reconstructed image and an actual image captured by the camera sensor. The reconstructed image is formed using the relative pose between consecutive RGB images and the estimated dense depth. The relative pose between the consecutive RGB images are computed using a pose estimation network~\cite{wong2021iccv} trained simultaneously with the depth completion network. The pose estimation network is used only during training and not deployed during testing. We follow the photometric loss defined in~\cite{jeon2022ral,wang2004tip}.

%The photometric loss is a combination of $L_1$ loss and structure similarity index measure~$\mathcal{SSIM}$~\cite{wang2004tip} and is defined as follows:
%\begin{equation}    
%\begin{aligned}
%l_p = \frac{1}{|\mathbf{\Omega}|} 
%\sum_{k \in \tau}^{}
%\sum_{\mathbf{p} \in \mathbf{\Omega}}^{} 
%w_1 \Vert \mathbf{I}_t(\mathbf{p}) - \hat{\mathbf{I}}_{k \rightarrow t}(\mathbf{p}) \Vert 
%\\
%+
%w_2 (1 - \mathcal{SSIM}({\mathbf{I}}_t(\mathbf{p}) - \hat{\mathbf{I}}_{k \rightarrow t}(\mathbf{p}))),
%\end{aligned}
%\end{equation}
%where~$\hat{\mathbf{I}}_{k \rightarrow t}$ denotes the reconstructed image from~$k$ to~$t$;~$\mathbf{p}$ and~$\mathbf{\Omega}$ denote single pixel and collection of pixel values in the image;~$w_1$ and~$w_2$ denote the weight of~$L_1$ loss and~$\mathcal{SSIM}$, respectively.

The sparse depth loss directly computes~$L_1$ distance between the estimated dense depth and the sparse depth input to prevent the loss of available information.
and is defined as follows:
\begin{equation}
\begin{aligned}
l_p = \frac{1}{|\mathbf{\Omega}_v|} \sum_{\mathbf{p} \in \Omega_v}^{} 
{(\Vert \mathbf{Z}_d(\mathbf{p}) - \mathbf{Z}_0(\mathbf{p}) \Vert)},
\end{aligned}
\end{equation}
where~$\mathbf{p}$ and $\mathbf{\Omega}_v$ denote single pixel and pixel values with valid sparse depth measurement, respectively.

The local smoothness loss ensures that the estimated dense depth has a smooth local gradient, preventing depth discontinuities between object boundaries and is defined as follows:
\begin{equation} 
\begin{aligned}
	l_s = \frac{1}{|\mathbf{\Omega}|} \sum_{\mathbf{p} \in \mathbf{\Omega}}^{} 
	{e^{-|\partial_x I(\mathbf{p})|} |\partial_x \mathbf{Z}_d(\mathbf{p})|
		+
		e^{-|\partial_y I(\mathbf{p})|} |\partial_y \mathbf{Z}_d(\mathbf{p})|},
\end{aligned}
%\vspace{-0.2cm}
\end{equation}
where $\mathbf{\Omega}$ denotes collection of pixel values in the image.

\begin{table}[t!]
	\centering
	\captionsetup{font=footnotesize}
	\caption{Error metric utilized for quantitative comparison of KITTI, NYUv2, and VOID datasets.~$\mathbf{Z}_{gt}\left(\cdot\right)$ denotes the ground truth depth.}
	%\resizebox{\columnwidth}{!}
	{\footnotesize 
		{
			\begin{tabular}{l|l}
				\toprule
				\midrule
				Error metric & Equation\\
				\midrule
				%Mean Absolute Error
				\textrm{MAE} & $\frac{1}{|\mathbf{\Omega}|}\sum_{\mathbf{p} \in \mathbf{\Omega}}\left\Vert \mathbf{Z}_d(\mathbf{p})-\mathbf{Z}_{gt}(\mathbf{p})\right\Vert$ \\
				%Root Mean-Square Error
				\textrm{RMSE} & $\left(\frac{1}{|\mathbf{\Omega}|}\sum_{\mathbf{p} \in \mathbf{\Omega}}\left\Vert \mathbf{Z}_d(\mathbf{p})-\mathbf{Z}_{gt}(\mathbf{p})\right\Vert_2\right)^{\frac{1}{2}}$\\
				%Inverse Mean Absolute Error
				\textrm{iMAE} &$\frac{1}{|\mathbf{\Omega}|}\sum_{\mathbf{p} \in \mathbf{\Omega}}\left\Vert 1/\mathbf{Z}_d(\mathbf{p})-1/\mathbf{Z}_{gt}(\mathbf{p})\right\Vert$\\
				%Inverse Root Mean-Square Error
				\textrm{iRMSE} & $\left(\frac{1}{|\mathbf{\Omega}|}\sum_{\mathbf{p} \in \mathbf{\Omega}}\left\Vert 1/\mathbf{Z}_d(\mathbf{p})-1/\mathbf{Z}_{gt}(\mathbf{p})\right\Vert_2\right)^{\frac{1}{2}}$\\
				\midrule
				\bottomrule
		\end{tabular}}
		\label{tab:error_metric}
	}
	\vspace{-0.5cm} % Adjust the vertical space as needed
\end{table}
%%%%%%%%%%%%%%%%%%%%%%%%%%%%%%%%%%%%%%%%%%%%%%%%%%%%%%%%%%%%%%%%%%%%%%%%%%%%%%%%
\section{Experimental Evaluation}
\label{sec:exp}

%% Repeat the main focus/objective with one single(!) sentence starting with:
%
%The main focus of this work is a  \dots

%% Explain the reader that the experiments with support all claims
%% (same list as in the intro!) starting the paragraph with:}
%
%We present our experiments to show the capabilities of our method. The results of our experiments also support our key claims, which are:
%
%(i)~\dots;
%
%(ii)~\dots;
%
%(iii)~\dots.

\subsection{Implementation Details}
We trained our network on a single NVIDIA RTX 3080 GPU with 12 GB memory. The batch size is set to 4 for both KITTI~\cite{uhrig2017threedv}, NYUv2~\cite{silberman2012eccv}, and VOID~\cite{wong2020ral} datasets. For the training, we adopt the Adam optimizer with exponential decay rates of~$\beta_1=0.9$ and $\beta_2=0.99$. For the KITTI dataset, the learning rate is scheduled as $5\times10^{-5}$, $1\times10^{-4}$, $1.5\times10^{-4}$, $1\times10^{-4}$, $5\times10^{-5}$, and $2\times10^{-5}$ for epochs 0-2, 3-8, 9-20, 21-30, 31-45, and 46-60, respectively. The image and sparse depth are center cropped to $768\times224$, removing areas with no sparse depth points, and the loss weights are set as $w_p=1$, $w_1=0.15$, $w_2=0.95$, $w_d=0.60$, and $w_l=0.06$. For the NYUv2 and VOID dataset, the learning rate is set to $1\times10^{-4}$ and $1.5\times10^{-4}$ for epochs 0-20 and 21-35, respectively. The image and sparse depth size are kept at $640\times480$ and loss weights are set as $w_p=1$, $w_1=0.15$, $w_2=0.95$, $w_d=1.0$, and $w_l=0.6$.

%We compare our method with other state-of-the-art unsupervised methods, such as KBNet~\cite{wong2021iccv}, VOICED~\cite{wong2020ral}, and ScaffNet-FusionNet~\cite{wong2021ral} on all datasets, using error metrics defined in~\mbox{\tabref{tab:error_metric}}. We also compared our method with \mbox{Struct-MDC}~\cite{jeon2022ral} which employs line features that correspond to a similar number of points on the NYUv2 and VOID dataset~(about 3,000~points).

While depth completion has seen recent progress, we focus on unsupervised methods suited for robotics applications where acquiring ground-truth depth is often impractical. To ensure fair evaluation, we compare our method with other well established unsupervised approaches that provide open-source implementations. Specifically, we evaluate against retrained KBNet~\cite{wong2021iccv}, VOICED~\cite{wong2020ral}, and \mbox{ScaffNet-FusionNet}~\cite{wong2021ral} using the error metrics defined in~\mbox{\tabref{tab:error_metric}}. Additionally, we compare our method with \mbox{Struct-MDC}~\cite{jeon2022ral} which employs line features and operates on a similar number of points in the NYUv2 and VOID datasets (about 3,000 points).
\begin{figure*}[t!]
	\captionsetup{font=footnotesize}
	\centering
	\def\svgwidth{0.95\textwidth}
	\graphicspath{{pics/0726_KITTI_compare/}} % important!
	\input{pics/0726_KITTI_compare/0726_KITTI_compare3.pdf_tex}
	\caption{Qualitative comparison on the KITTI depth completion dataset. (a)~Input RGB image, (d)~LiDAR sparse depth, and estimated depth results from (b)~VOICED~\cite{wong2020ral}, (c)~FusionNet~\cite{wong2021ral}, (e)~KBNet~\cite{wong2021iccv}, and (f)~ours. The white and cyan rectangles denote the area where our network can distinguish between an object and its background compared with other networks.}
	%USE INKSCAPE!!!! It's a perfect drawing tool for Linux users. Please set \texttt{\textbackslash captionsetup$\{$font$=$footnotesize$\}$} and \texttt{\textbackslash vsfig}. These significantly reduce some spaces and make the paper more luxurious haha~(best viewed in color).}
	\label{fig:KITTI_compare}
	\vspace{-0.5cm}
\end{figure*}

\subsection{KITTI Dataset Evaluation}
\begin{table}[t!]
	\centering
	\captionsetup{font=footnotesize}
	\caption{Quantitative comparison results on the KITTI dataset. The bold and underline denote the best and second-best performance. (Units for MAE and RMSE: mm, iMAE and iRMSE: 1/km)}
	{\footnotesize
		{
			\begin{tabular}{l|cccc}
				\toprule
				\midrule
				KITTI & MAE & RMSE & iMAE & iRMSE\\
				\midrule
				KBNet~\cite{wong2021iccv} & \bftab{296.810} & \underline{1174.773} & \bftab{1.101} & \underline{3.130} \\
				VOICED~\cite{wong2020ral} & 311.276 & 1247.049 & 1.380 & 3.354 \\
				FusionNet~\cite{wong2021ral} & 305.392 & 1216.522 & 1.304 & 3.527 \\
				\midrule
				Ours & \underline{303.090} & \bftab{1159.027} & \underline{1.134} & \bftab{2.976} \\
				\midrule
				\bottomrule
			\end{tabular}
			\label{table:result_KITTI}
		}
		\vspace{-0.3cm}
	}
\end{table}
%\begin{table}[t!]
%	\centering
%	\captionsetup{font=footnotesize}
%	\caption{Quantitative comparison results on the KITTI dataset. The bold and underline denote the best and second-best performance. (Units for MAE and RMSE: mm, iMAE and iRMSE: 1/km)}
%	{\footnotesize
%		{
%			\begin{tabular}{l|cccc}
%				\toprule
%				\midrule
%				KITTI & MAE & RMSE & iMAE & iRMSE\\
%				\midrule
%				KBNet$^{4}$ & \bftab{296.810} & \underline{1174.773} & \bftab{1.101} & \underline{3.130} \\
%				VOICED$^{5}$ & 311.276 & 1247.049 & 1.380 & 3.354 \\
%				FusionNet$^{6}$ & 305.392 & 1216.522 & 1.304 & 3.527 \\
%				\midrule
%				Ours & \underline{303.090} & \bftab{1159.027} & \underline{1.134} & \bftab{2.976} \\
%				\midrule
%				\bottomrule
%			\end{tabular}
%			\label{table:result_KITTI}
%		}
%		\vspace{-0.3cm}
%	}
%\end{table}

\begin{table}[t!]
	\centering
	\captionsetup{font=footnotesize}
	\caption{Quantitative comparison results on the NYUv2 dataset. The bold and underline denote the best and second-best performance. (Units for MAE and RMSE: mm, iMAE and iRMSE: 1/km)}
	{\footnotesize
		{
			\begin{tabular}{l|cccc}
				\toprule
				\midrule
				NYUv2 & MAE & RMSE & iMAE & iRMSE\\
				\midrule
				KBNet~\cite{wong2021iccv} & 41.421 & 90.161 & \bftab{7.118} & 20.907 \\
				VOICED~\cite{wong2020ral} & 68.458 & 134.097 & 12.581 & 25.151 \\
				FusionNet~\cite{wong2021ral} & 67.082 & 126.967 & 12.521 & 24.981 \\
				Struct-MDC~\cite{jeon2022ral} & \bftab{36.968} & \underline{85.818} & 7.853 & \underline{19.500} \\
				\midrule 
				Ours & \underline{38.125} & \bftab{80.063} & \underline{7.280} & \bftab{16.094} \\
				\midrule
				\bottomrule
		\end{tabular}}
		\label{table:result_NYUv2}
	}
	\vspace{-0.6cm}
\end{table}
As shown in~\figref{fig:KITTI_compare}, other networks do not clearly show distinguishing boundaries between objects and their backgrounds. The estimated depth map incorrectly predicts the background as a part of the object due to ambiguous differences in the extracted image feature. This result is due to their overreliance on convolution layers that smooth out depth and image features in the encoder and decoder layers. Our network can correctly identify the boundaries between the background and objects, particularly for objects far from the sensor setup.
%Our network can correctly identify the boundaries between the background and objects. This advantage is particularly evident for objects far from the sensor setup. 
For example, the cars on the first and second rows in~\figref{fig:KITTI_compare} are clearly distinguished in the estimated dense depth map, even though the sparse depth input does not fully cover the car. On the third row, our results are more dense and complete compared with other network results that have holes on the car. This result is due to transformer decoder blocks in our network separating the background and object from the smoothed encoder features by focusing on different feature parts through window partitions. 

Moreover, the \mbox{cross-hierarchical-attention} module successfully refines the estimated dense depth map, leading to a quantitatively more accurate output, as demonstrated in~\tabref{table:result_KITTI}. \mbox{Cross-hierarchical-attention} has been shown to enhance accuracy by leveraging information from multiple sources, reducing large errors in the depth map, as seen in RMSE and iRMSE improvement, while retaining similar MAE and iMAE. Our network reduces the error by an average of 2\% and up to 5\% on the iRMSE metric compared with the baseline. This improvement confirms our network's ability to clearly distinguish details between objects and their background, resulting in a higher accuracy.

\subsection{NYUv2 Dataset Evaluation}
In contrast to the KITTI dataset, NYUv2 provides measurements in an indoor environment.~\figref{fig:nyuv2_compare} compares our network with other networks using simulated LiDAR depth measurements. In other networks, some objects merge with other items or backgrounds, resulting from the network's failure to extract distinguishing features between different entities. However, our network successfully produces complete depth maps that retain detailed edges between objects. This success can be attributed again to our transformer architecture which is able to focus on detailed parts of the feature compared with other networks. Despite having lower parameters and runtime requirements, our network's result is more accurate, especially in separating objects, as seen in an average of 9\% improvement to both MAE and RMSE compared with the baseline, as shown in~\tabref{table:result_NYUv2}. The MAE result of Struct-MDC is higher than ours due to the addition of line features that give more information compared with sparse depth points.

\begin{figure*}[t!]
	\captionsetup{font=footnotesize}
	\centering
	\def\svgwidth{0.91\textwidth}
	\graphicspath{{pics/0727_nyuv2_compare/}} % important!
	\input{pics/0727_nyuv2_compare/0727_nyuv2_compare3.pdf_tex}
	\caption{Qualitative comparison on the NYUv2 dataset. (a)~Input RGB image, (b)~VOICED~\cite{wong2020ral}, (c)~FusionNet~\cite{wong2021ral}, (d)~KBNet~\cite{wong2021iccv}, (e)~ours, and (f)~ground truth. The dashed rectangles denote the area where our network can produce notable object outlines compared with other networks, as seen in the sharper contrasting color similar to the ground truth.}
	%USE INKSCAPE!!!! It's a perfect drawing tool for Linux users. Please set \texttt{\textbackslash captionsetup$\{$font$=$footnotesize$\}$} and \texttt{\textbackslash vsfig}. These significantly reduce some spaces and make the paper more luxurious haha~(best viewed in color).}
	\label{fig:nyuv2_compare}
	\vspace{-0.5cm}
\end{figure*}

\subsection{VOID dataset evaluation}
\begin{figure*}[t!]
	\captionsetup{font=footnotesize}
	\centering
	\def\svgwidth{0.91\textwidth}
	\graphicspath{{pics/0219_void_compare/}} % important!
	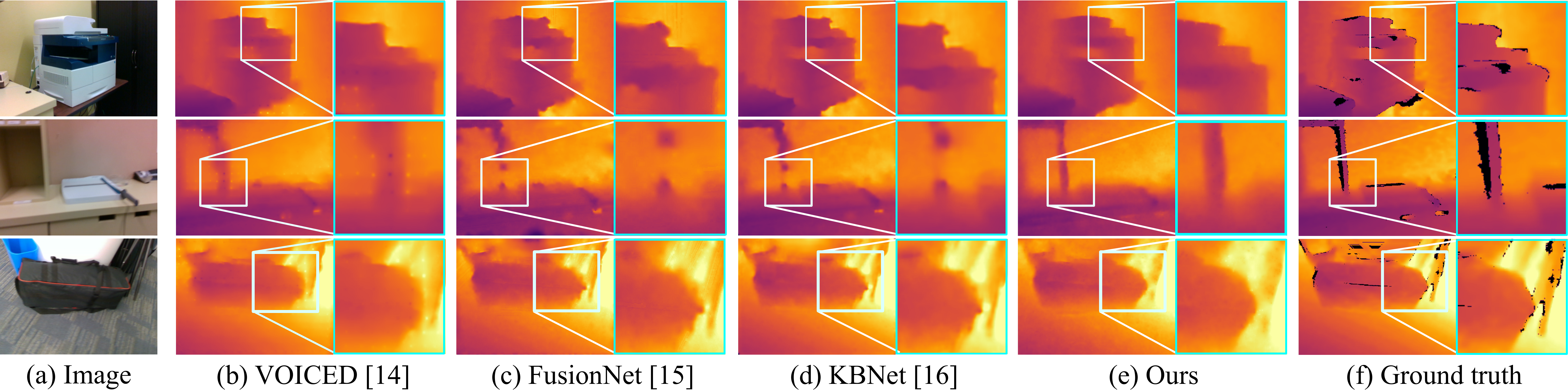
	\caption{Qualitative comparison on the VOID dataset. (a)~Input RGB image, (b)~VOICED~\cite{wong2020ral}, (c)~FusionNet~\cite{wong2021ral}, (d)~KBNet~\cite{wong2021iccv}, (e)~ours, and (f)~ground truth. The white and cyan rectangles denote the area where our network can produce notable object outlines compared with other networks, as seen in the sharper contrasting color similar to the ground truth.}
	%USE INKSCAPE!!!! It's a perfect drawing tool for Linux users. Please set \texttt{\textbackslash captionsetup$\{$font$=$footnotesize$\}$} and \texttt{\textbackslash vsfig}. These significantly reduce some spaces and make the paper more luxurious haha~(best viewed in color).}
	\label{fig:void_compare}
	\vspace{-0.6cm}
\end{figure*}

The VOID dataset also provides depth measurements in various indoor environments at closer distances.~\tabref{table:result_VOID} shows that KBNet and Struct-MDC achieve lower MAE and iMAE than other networks. This is due to their feature extraction modules which utilize denser convolution layers and additional line features, respectively. However, both networks suffer from higher RMSE and iRMSE due to outliers in the predicted depth maps.  In contrast, our network demonstrates its stability by achieving the lowest RMSE and iRMSE, with an improvement of 10\% compared with its baseline, among all methods. This improvement can be attributed to our \mbox{cross-hierarchical-attention} module which effectively compares RGB and depth features to reduce outliers in the predicted depth maps.~\figref{fig:void_compare} shows the qualitative comparison between our network and other methods using simulated LiDAR depth measurements. In the first and third rows, other networks struggle to maintain object outlines. Conversely, our network successfully preserves object shapes by leveraging RGB features to enhance shape definition. This issue is even more apparent in the second row. Our network is the only one that accurately reproduces object edges while other methods incorrectly predict holes in the object edges. These results further confirm our network's
ability to distinguish objects and their backgrounds clearly. 

\begin{table}[t!]
	\centering
	\captionsetup{font=footnotesize}
	\caption{Quantitative comparison results on the VOID dataset. The bold and underline denote the best and second-best performance. (Units for MAE and RMSE: mm, iMAE and iRMSE: 1/km)}
	{\footnotesize
		{
			\begin{tabular}{l|cccc}
				\toprule
				\midrule
				VOID & MAE & RMSE & iMAE & iRMSE\\
				\midrule
				KBNet~\cite{wong2021iccv} & \underline{47.618} & 105.395 & 25.244 & 57.553 \\
				VOICED~\cite{wong2020ral} & 54.116 & 114.972 & 25.719 & \underline{45.436} \\
				FusionNet~\cite{wong2021ral} & 53.119 & 114.751 & 31.909 & 75.022 \\
				Struct-MDC~\cite{jeon2022ral} & \bftab{30.680} & \underline{98.692} & \bftab{14.303} & 59.121 \\
				\midrule 
				Ours & 48.982 & \bftab{95.067} & \underline{22.034} & \bftab{36.608} \\
				\midrule
				\bottomrule
		\end{tabular}}
		\label{table:result_VOID}
	}
	\vspace{-0.5cm}
\end{table}

\subsection{Ablation Study}
An ablation study was performed on the channel and window sizes, which significantly impact the number of parameters. We designed the network to minimize memory usage while achieving the performance needed to surpass the baseline. Additionally, we conducted experiments using the image feature as query, instead of the depth feature. As shown in~\tabref{table:ablation_study}, our network performed best with the depth feature as the query. 
We hypothesize that this result is due to the image feature, as the key-value pair, preserving more edges than the depth feature.
Additionally, 
%The network's performance dropped with larger channel and window sizes, likely due to underfitting from lacking transformer pretraining to compensate for the limited training data. However, our network achieves strong results without pretraining despite limited training data.
the network's performance was decreased across all metrics as channel and window sizes were increased, likely due to underfitting, as transformers typically require pretraining to compensate for limited data. Despite limited data and without pretraining, our network achieves strong results with minimal parameters.
%It is also observed that our network performed worse on all metrics as the channel and window size increased. We believe this is due to the underfitting phenomenon, as transformers are notoriously known to perform worse without pre-training. However, our network is able to obtain a satisfying result despite the lack of sufficient training data with the lowest amount of network parameters.

\begin{table}[t!]
	\centering
	\captionsetup{font=footnotesize}
	\caption{Ablation study of our network on query type, channel size, and window size on the NYUv2 dataset. The bold and underline denote the best and second-best performance. (Units for MAE and RMSE: mm, iMAE and iRMSE: 1/km)}
	{\footnotesize \resizebox{\columnwidth}{!}
		{
			\begin{tabular}{l|c|c|ccccc}
				\toprule
				\midrule
				Query & Channel size & Window size & MAE & RMSE & iMAE & iRMSE & params \\
				\midrule
				{\multirow{4}{*}{Depth}} & [16,\ 32,\ 64,\ 128] & [2,\ 2,\ 4,\ 4] & \bftab{38.125} &    \bftab{80.063} & \bftab{7.280} & \bftab{16.094}   
				& \bftab{1.1M} \\
				& [16,\ 32,\ 64,\ 128] & [4,\ 4,\ 8,\ 8] & 53.508 & 110.590 & 11.067 & 29.657 & \underline{1.6M} \\
				& [32,\ 64,\ 128,\ 256] & [2,\ 2,\ 4,\ 4] & 73.940 & 121.003 & 13.634 & 28.210 & 3.7M\\
				& [32,\ 64,\ 128,\ 256] & [4,\ 4,\ 8,\ 8] & 46.104 & 91.803 & 8.041 & \underline{17.503} & 4.2M \\
				\midrule
				{\multirow{4}{*}{Image}} & [16,\ 32,\ 64,\ 128] & [2,\ 2,\ 4,\ 4] & \underline{39.724} & \underline{85.242} & \underline{7.968} & 18.334 & \bftab{1.1M} \\
				& [16,\ 32,\ 64,\ 128] & [4,\ 4,\ 8,\ 8] & 51.564 & 104.066 & 11.806 & 27.626 & \underline{1.6M}\\
				& [32,\ 64,\ 128,\ 256] & [2,\ 2,\ 4,\ 4] & 54.064 & 98.498 & 14.098 & 27.990 & 3.7M\\
				& [32,\ 64,\ 128,\ 256] & [4,\ 4,\ 8,\ 8] & 45.761 & 95.101 & 8.900 & 17.844 & 4.2M \\
				\midrule
				\bottomrule
		\end{tabular}}
		\label{table:ablation_study}
		\vspace{-0.5cm}
	}
\end{table}

\subsection{Memory \& Computation Analysis}
The number of model parameters and average runtimes of the algorithms are presented in \tabref{table:model_param_runtime}. Our network consumes smaller memory than other networks, with 5.3M parameter less than the baseline, KBNet. Moreover, in terms of runtime, our network takes only 11.5~ms~(86.96~Hz) compared with the baseline taking 13.56~ms~(73.74~Hz). This improvement can be attributed to our careful design of the encoder and decoder layers using the depthwise convolutional layers and squeeze-and-excite modules. A depthwise convolutional layer only needs one filter per input channel and outputs the same number of channels as input, resulting in fewer parameters per filter. The channel information can also be preserved using the squeeze-and-excite module, improving channel interaction even if the previous depthwise convolutional layers did not. This result demonstrates our network's computational efficiency and~\mbox{real-time} capability to predict the complete depth map without sacrificing accuracy.    

\section{Conclusion}
\label{sec:conclusion}
In this paper, we proposed {CHADET} that can combine an RGB image with sparse depth points to produce a complete depth map. We successfully improved the computational efficiency of our network by combining a depthwise convolutional layer and squeeze-and-excite module. In terms of accuracy, the full \mbox{cross-hierarchical-attention} module is able to distinguish the difference between objects and backgrounds during the learning process. Thus, {CHADET} outputs a depth map with lower error compared with other networks. For future work, the network could be enhanced to handle larger environmental variations by incorporating a more sophisticated loss function, such as the contrastive loss. The attention mechanism computes the similarity scores between the query and key features, which can be incorporated into the contrastive loss through careful design considerations to improve the learning process. Additionally, the depthwise block can also utilize a \mbox{self-attention} mechanism that focuses on distinguishing objects.

\vspace{-0.2cm}
\begin{table}[t!]
	\centering
	\captionsetup{font=footnotesize}
	\caption{Model parameters size and average runtime.}
	{\footnotesize{}
		{
			\begin{tabular}{l|ccccc}
				\toprule
				\midrule
				Model & Params & Runtime~(ms) \\
				\midrule
				KBNet~\cite{wong2021iccv} & 6.4M & {13.56} \\
				VOICED~\cite{wong2020ral} & 10M & {18.53}\\
				FusionNet~\cite{wong2021ral} & 6.7M & {17.56} \\
				Ours & \bftab{1.1\textrm{M}} & \bftab{11.5} \\
				\midrule
				\bottomrule
		\end{tabular}}
		\label{table:model_param_runtime}
	}
	\vspace{-0.6cm}
\end{table}

%Future works could enhance the algorithm by using more robust triangulation methods that consider UWB's high-rate information. Furthermore, a deep learning framework that estimates the error can also be incorporated for better optimization.
%In this paper, we presented a novel approach to\dots
%Our approach operates \dots  Our method exploits \dots
%This allows us to successfully \dots
%We implemented and evaluated our approach on different datasets
%and provided comparisons to other existing techniques and supported
%all claims made in this paper. The experiments suggest that \dots

%%%%%%%%%%%%%%%%%%%%%%%%%%%%%%%%%%%%%%%%%%%%%%%%%%%%%%%%%%%%%%%%%%%%%%%%%%%%%%%%
%% Future work: Use only if applicable -- but if so, use the following
%% sentence to start:
% Despite these encouraging results, there is further space for improvements. 

%%%%%%%%%%%%%%%%%%%%%%%%%%%%%%%%%%%%%%%%%%%%%%%%%%%%%%%%%%%%%%%%%%%%%%%%%%%%%%%%
% Only if applicable
%\section*{Acknowledgments}
%We thank XXX for fruitful discussions and for \dots

\bibliographystyle{URL-IEEEtrans}

% All new citations should go to new.bib. The file glorified.bib should go
% be the one from the ipb server. After paper or related work has been
% written merge the entries from new.bib to glorified.bib ON THE SERVER,
% replace the glorified.bib in this repository and empty the new.bib
\bibliography{URL-bib}

\end{document}